\documentclass[electronics,article,accept,moreauthors]{Definitions/mdpi} 
\firstpage{1} 
\pubvolume{1}
\issuenum{1}
\articlenumber{0}
\pubyear{2026}
\copyrightyear{2026}
\externaleditor{Hector E. Nistazakis} % More than 1 editor, please add `` and '' before the last editor name
\datereceived{22 January 2026} 
\daterevised{15 February 2026} % Comment out if no revised date
\dateaccepted{24 February 2026} 
\datepublished{ } 
\usepackage{algorithm}
\usepackage{algorithmic}
\usepackage{xcolor}
\Title{PPO-Based %MDPI: Notes for Authors
 Hybrid Optimization for RIS-Assisted Semantic Vehicular Edge Computing} %Attention: Title changed.

\Author{Wei Feng %MDPI: Please carefully check the accuracy of names and affiliations. Please notice that any changes about authorship, affiliations, and corresponding authors are not allowed after the paper has been accepted, including adding, removing, or changing the order. No further modification is allowed after confirmation.
 $^{1,\dagger}$, Jingbo Zhang $^{1,\dagger}$, Qiong Wu $^{1,}$*, Pingyi Fan $^{2}$ and Qiang Fan $^{3}$}

\AuthorNames{Wei Feng, Jingbo Zhang, Qiong Wu, Pingyi Fan and Qiang Fan}

\address{%
$^{1}$ \quad School of Internet of Things Engineering, Jiangnan University, Wuxi 214122, China; \linebreak  fengwei@jiangnan.edu.cn (W.F.); jingbozhang@stu.jiangnan.edu.cn (J.Z.)\\

$^{2}$ \quad State Key Laboratory %MDPI: Please arrange the authors’ address information from subordinate to superior. 
 of Space Network and Communications, Beijing National Research Center for Information Science and Technology, Department of Electronic Engineering, %MDPI: Please confirm if ``and the'' is necessary; if not, please remove it. 
  Tsinghua University, \mbox{Beijing 100084,} China; fpy@tsinghua.edu.cn\\

$^{3}$ \quad Qualcomm, San Jose, CA 95110, USA; qianfan@qti.qualcomm.com %MDPI: The email address highlighted is different from the ones submitted online at susy.mdpi.com (qianfan@qti.qualcomm.com). Please confirm which are correct.
}
\corres{Correspondence: qiongwu@jiangnan.edu.cn}
\firstnote{These authors contributed equally to this work.}
\abstract{To support latency-sensitive Internet of Vehicles (IoV) applications amidst dynamic environments and intermittent links, this paper proposes a Reconfigurable Intelligent Surface (RIS)-aided semantic-aware Vehicle Edge Computing (VEC) framework. This approach integrates RIS to optimize wireless connectivity and semantic communication to minimize latency by transmitting semantic features. We formulate a comprehensive joint optimization problem by optimizing offloading ratios, the number of semantic symbols, and RIS phase shifts. Considering the problem's high dimensionality and non-convexity, we propose a two-tier hybrid scheme that employs Proximal Policy Optimization (PPO) for discrete decision-making and Linear Programming (LP) for offloading optimization. {The simulation results have validated the proposed framework's superiority over existing methods. Specifically, the proposed PPO-based hybrid optimization scheme reduces the average end-to-end latency by approximately 40\% to 50\% compared to Genetic Algorithm (GA) and Quantum-behaved Particle Swarm Optimization (QPSO). Moreover, the system demonstrates strong scalability by maintaining low latency even in congested scenarios with up to 30 vehicles.}}

\keyword{vehicular %MDPI: we revised all the first letters to be lowercase by our standard, please confirm.
 edge computing; reconfigurable intelligent surface; semantic \linebreak    communication; task offloading} 

\begin{document}

%%%%%%%%%%%%%%%%%%%%%%%%%%%%%%%%%%%%%%%%%%

\section{Introduction %MDPI: Dear authors,
%1. If the use of software is mentioned in the text, Please state the version number of the software in the whole text wherever they are mentioned first. If no version number, please add available website link and accessed date.
%2. If the chemicals & reagents, devices, instruments, commercial cell lines/samples/materials used in the article, please state the name of the manufacturer, city, and country from where the equipment was sourced.
%3. If the standard used in the article, ref citations should be added for Standard (at least in the first-mentioned position), please add the refs to the References list and cite it after standard. Please provide the detailed information such as: Standard’s Number; Standard’s Title. Publisher: City, Country, Year. 
% I have confirmed this part accordingly.
} \label{sect:s1}
The Internet of Vehicles (IoV) is driving the intelligent transformation of transportation systems by enabling real-time connectivity between vehicles, infrastructure, and~pedestrians to support autonomous driving. However, the~massive high-frequency data generated by onboard sensors is challenging in traditional central cloud architectures, especially to meet low-latency and high-reliability requirements~\cite{ref4,ref5,ref6,refZheng2021,refWu2023,refChen2017,refYang2012,refZhang2008,refFan2002,refFan1999,refZhou2011,refWang2010,refYao2003}. To~address this, Vehicular Edge Computing (VEC) allows vehicles to offload tasks to edge servers (ESs) at roadside units (RSUs), which significantly reduces latency and energy consumption by shortening communication distances~\cite{ref7}, effectively alleviating core network~loads.

Nevertheless, urban obstacles and multipath fading often degrade vehicle--RSU links~\cite{ref10}, potentially violating Quality of Service (QoS) standards~\cite{ref11}. Thus, reconfigurable Intelligent Surfaces (RISs) offer a solution by dynamically adjusting electromagnetic signal properties to reroute signals, boost strength, and~suppress interference~\cite{ref13}. Research on RIS-aided V2X has optimized these critical links~\cite{ref17,ref18}. {As investigated in~\cite{ref48}, RIS empowers networks to proactively manipulate the wireless channel rather than passively adapting to it. Specifically, each passive element on the RIS can independently impose a controlled phase shift on the incident electromagnetic waves. By~optimizing the phase-shift matrix, RIS reshapes the propagation environment to compensate for high-frequency attenuation and urban shadows. This mechanism is particularly crucial for semantic-aware VEC, as~it ensures high-fidelity transmission of semantic features which are highly sensitive to channel quality.}

{Concurrently, semantic communication has emerged as a paradigm shift by moving beyond traditional Shannon entropy limits to focus on the effective transmission of meaning. A~foundational DeepSC framework proposed by~\cite{ref57} has employed deep learning to jointly optimize source and channel encoding. Unlike bit-level transmission, it extracts and transmits only task-relevant semantic features, thereby reducing data volume while maintaining high information integrity even in a low signal-to-interference-plus-noise ratio (SINR) environments.} This approach ensures that minor decoding errors do not compromise information integrity~\cite{ref24}, while the reduced data volume significantly lowers latency~\cite{ref25}.

Integrating semantic communication with RIS-aided VEC becomes a critical development direction. RIS enhances signal coverage against attenuation, while semantic communication minimizes transmission overhead and enables intelligent resource scheduling~\cite{ref26,ref27}, thus delivering high adaptability under extreme conditions~\cite{ref28}. However, high vehicle mobility causes rapid channel variations, making traditional static optimization methods ineffective for such non-convex problems. Consequently, artificial intelligence~\cite{ref29}, specifically deep reinforcement learning (DRL) \cite{ref35}, has gained prominence for its ability to solve complex dynamic~problems.

The integration of these components presents several challenges, including complex semantic extraction, time-varying link quality, and~difficulty of jointly optimizing offloading ratios, RIS phase shifts, and~the number of semantic symbols. To~address these issues, this paper proposes a novel optimization framework for an RIS-aided, semantic-aware VEC system. The~Proximal Policy Optimization (PPO) algorithm is employed to optimize RIS phase shifts and the number of semantic symbols, while Linear Programming (LP) is used to allocate offloading ratios. The~main contributions are summarized as follows:

\begin{enumerate}[label=$\bullet$]
	\item We propose an RIS-aided semantic-aware VEC system, in~which a three-path semantic task partitioning framework is designed. Each vehicular task is adaptively divided into three parts: local execution, a~vehicle-to-infrastructure (V2I) task offloaded to the RSU, and~a vehicle-to-vehicle (V2V) task offloaded to a service vehicle (SV). A~link-level RIS enhancement mechanism is introduced, enabling the RIS to improve the quality of either the V2I and V2V link based on real-time semantic similarity, mobility, and~channel conditions. 
	\item We formulate a comprehensive joint optimization problem involving offloading rates, the~number of semantic symbols, and~RIS phase shifts in~order to minimize transmission latency. To~address the non-convexity of this problem, we propose a two-layer collaborative hybrid framework: PPO is employed to make discrete decisions on the number of semantic symbols and RIS phase shifts, while LP is utilized to optimize the offloading ratio.
	\item Extensive simulation results have demonstrated that the proposed method outperforms other traditional schemes significantly in terms of  end-to-end latency.
\end{enumerate}

{The remainder of this paper is organized as follows: Section~\ref{sect:s2} reviews the related work on DRL-based task offloading, RIS-aided VEC systems, and~the latest advancements in semantic communications. Section~\ref{sect:s3} describes the RIS-aided semantic-aware VEC system model and formulates the joint optimization problem for minimizing system latency. Section~\ref{sect:s4} details the proposed two-layer collaborative hybrid framework, which integrates the PPO algorithm for discrete configuration and LP for continuous task offloading. The~computational complexity of the proposed scheme is also analyzed in this section. \mbox{Section~\ref{sect:s5}} presents the simulation setup and discusses the performance results of the proposed method in comparison with several benchmarks. Finally, Section~\ref{sect:s6} concludes this paper and discusses future research directions.}
\section{Related~Work} \label{sect:s2}
In this section, we first review DRL-based Mobile Edge Computing (MEC) studies, then summarize existing research on RIS-aided VEC task offloading, and~finally discuss semantic communication-assisted VEC~work.

Several studies have applied DRL to VEC~\cite{ref_gu2025drl}, addressing critical issues such as pricing, security~\cite{ref_xiao2026vehicular}, and~task dependencies~\cite{ref_chu2026cavs,refWu2015,refWu2014,refWu2014b,refFan2010,refGu2025}. Wang~et~al.~\cite{ref36} proposed an Actor-Critic-based task offloading model that utilizes a partial offloading strategy to effectively reduce system costs in multi-vehicle scenarios. Their approach accelerates policy convergence, offering a novel solution for real-time VEC optimization. From~an incentive mechanism's perspective, Wu~et~al.~\cite{ref37} introduced a pricing-driven framework and developed Pricing-Driven Resource Allocation (PDRA) and DRL-based Pricing-Driven Dynamic Resource Allocation (DPDDRA) algorithms. These methods enable scalable and efficient resource management adaptable to rapid mobility and channel variations. Addressing tasks with strong dependencies and strict latency constraints, Zeng~et~al.~\cite{ref38} proposed a joint optimization method using Double Deep Q-Network (DDQN). By~comprehensively modeling task priorities and dependencies, they achieve collaborative optimization of offloading decisions, bandwidth, and~computing resources within a multi-stage decision-making process. Furthermore, Ren~et~al.~\cite{ref39} integrated blockchain and Asynchronous Advantage Actor-Critic (A3C) into the VEC architecture. Through a two-layer Software-Defined Network (SDN) %MDPI: Please confirm if it is a software. If yes, please state which version of the software was used.
% Software-Defined Network (SDN) is a network architecture rather than a specific software product. Therefore, it does not have a version number or manufacturer. In this context, we refer to the general SDN framework as discussed in the cited literature.
 comprising domain and regional controls, they realized secure and trusted service migration, significantly reducing latency and energy consumption while improving throughput. In~summary, while these DRL-based studies demonstrate significant adaptability, they predominantly rely on traditional communication links, largely neglecting the potential of RIS-aided environments and semantic-aware~models.

The integration of RIS in VEC has been explored for energy efficiency~\cite{ref47, ref48} and security~\cite{ref49}, yet often neglects semantic aspects. Wang~et~al.~\cite{ref43} proposed an RIS-aided NOMA VEC architecture, employing a two-stage offloading scheme to effectively reduce vehicle energy consumption. Addressing coverage blind spots, Xie~et~al.~\cite{ref44} utilized dual-roadside RISs and achieved near-optimal real-time offloading via multi-dimensional parameter optimization. In~aerial-assisted networks, Rzig~et~al.~\cite{ref45} developed a HAPS-UAV cooperative framework, using iterative algorithms to balance offloading success rates and energy usage. Similarly, Xiao~et~al.~\cite{ref46} maximized the minimum weighted throughput in an RIS-UAV MEC system through a synchronous offloading scheme. Collectively, while validating the benefits of RIS in VEC, these works remain confined to traditional links, lacking integration with semantic communication and dynamic link-level RIS~selection.

Recent studies have significantly advanced semantic VEC through AI-driven mechanisms, covering areas from efficient offloading~\cite{ref54} and privacy-preserving frameworks~\cite{ref55} to robust transmission in complicated conditions. Specifically, Yang~et~al.~\cite{ref50} utilized a diffusion-based multi-agent reinforcement learning algorithm to enhance exploration capabilities, effectively reducing system latency and costs. To~improve transmission reliability, Du~et~al.~\cite{ref51} integrated Large Multimodal Models (LMMs) into vehicular networks, enhancing task accuracy in low-SINR environments through task-oriented semantic mechanisms. Addressing model maintenance, Zheng~et~al.~\cite{ref52} proposed a privacy-preserving personalized federated learning framework to overcome privacy and mobility bottlenecks. Furthermore, Chen~et~al.~\cite{ref53} developed a semantic-aware multimodal model using an MAPPO algorithm to jointly optimize offloading and resource allocation, thereby boosting the Quality of Experience (QoE).

Despite these advancements, existing research often optimizes semantic extraction and resource allocation in isolation. But~there are few comprehensive studies that jointly model semantic communication with physical link characteristics, particularly neglecting synergy between RIS assistance and multipath offloading strategies. {Mao~et~al.~\cite{refmao} proposed an RIS-enhanced semantic-aware sensing, communication, computation, and~control (SC3) network framework, focusing on closed-loop control optimization in industrial IoT scenarios. In~contrast, this paper focuses on the highly dynamic scenarios of the IoV, proposing a unique three-path semantic task partitioning framework, and~specifically models the randomness introduced by vehicle mobility, which is fundamentally different from the fixed-device scheduling in industrial scenarios.} {Zhao~et~al.~\cite{refzhao} investigated traffic reshaping and channel reconfiguration in RIS-aided semantic NOMA networks, with~an emphasis on optimizing decoding order to reduce energy consumption. In~comparison, this paper introduces V2V cooperative semantic offloading while leveraging SVs as auxiliary computing nodes, which makes more efficient use of idle computing resources in the IoV than pure access point (AP) offloading.} {Xie~et~al.~\cite{refxie} studied the fairness of semantic communication systems assisted by STAR-RIS in order to enhance semantic similarity by maximizing the minimum SINR. Different from their work, this paper optimizes the physical-layer channel via RIS, but~also directly takes the number of semantic symbols as a decision variable for cross-layer optimization. This three-dimensional joint optimization scheme, which deeply couples the physical-layer phase matrix with the application-layer semantic compression depth, offers greater flexibility than beamforming alone when handling latency-sensitive~tasks.}

To the best of our knowledge, this is among the first works to jointly consider offloading allocation, RIS phase shift optimization, and~the number of semantic symbols in an RIS-aided semantic-aware VEC~system.

\section{System Model and Problem~Formulation} \label{sect:s3}
\begin{figure}[H]
	 
	\includegraphics[scale=0.4]{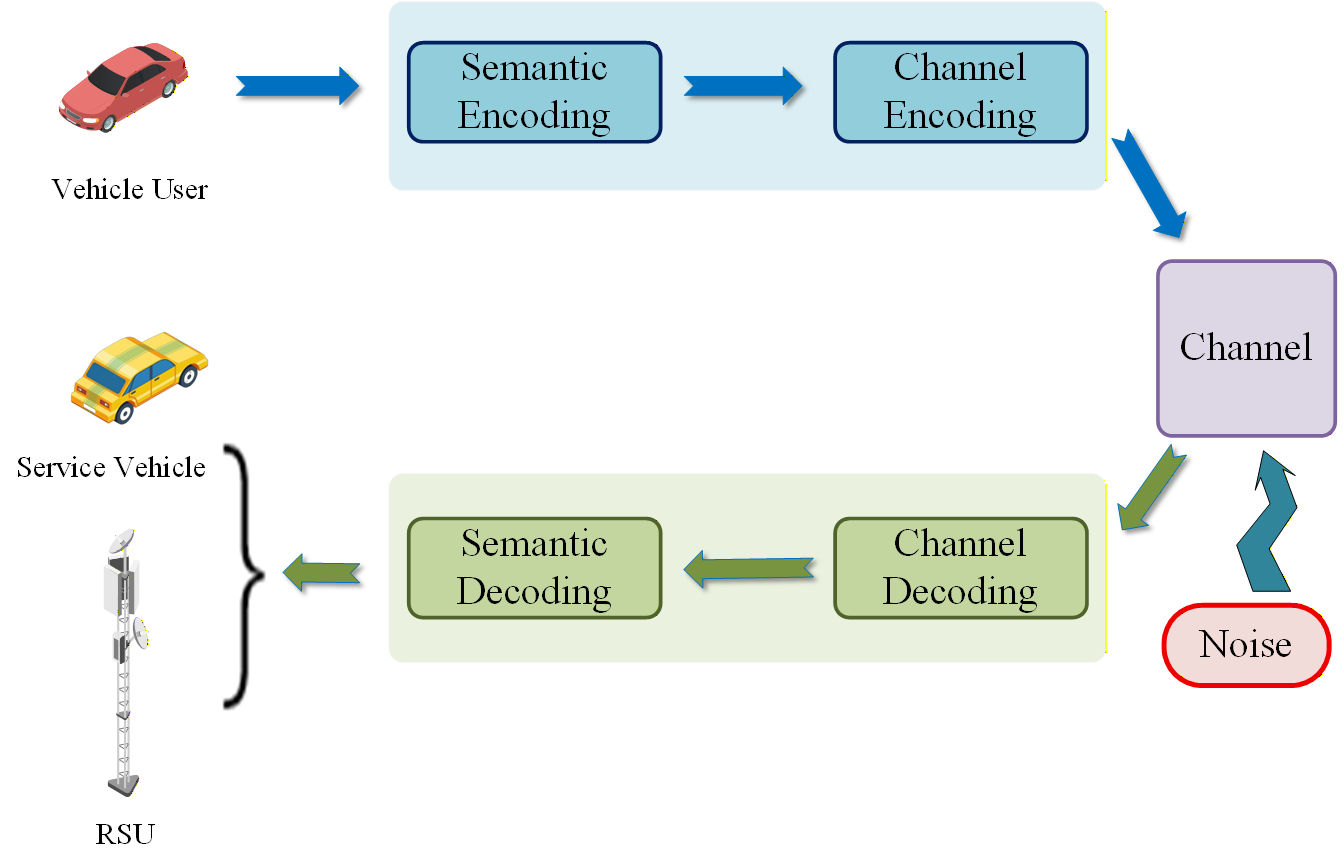}
	\caption{Framework of semantic communication~system.}
	\label{fig:figsemantic}
\end{figure}

In recent years, with~the continuous growth of wireless communication demands, a~small bit error in traditional communication systems may lead to a significant semantic error. In~contrast, in~semantic communication, errors from the decoding process do not affect the integrity of the transmitted information. For~example, the~words ``youth'' and ``Adolescent'' can be mapped to the same semantic symbols, even though their spellings are completely different. Therefore, we utilize the pre-trained DeepSC model~\cite{ref57} to improve communication efficiency by focusing on the meaning of the transmitted data rather than just the raw data itself. The~semantic communication framework is illustrated in {Figure} \ref{fig:figsemantic}, which consists of semantic encoding, channel encoding, channel decoding, semantic decoding, the~channel, and~noise. All vehicles and RSUs share the same semantic knowledge base and are equipped with the DeepSC system. {However, it is important to note that maintaining knowledge base consistency in highly dynamic IoV environments presents a significant challenge. In~this study, we focus on the joint optimization of resource allocation under a static knowledge base as a foundational step. The~potential overhead and latency associated with real-time, dynamic knowledge base updates—necessitated by varying semantic tasks—are considered a limitation of the current framework and will be explored in future research.} We assume that the transmitted task features are consistent with those in~\cite{ref57}, and~thus directly reuse the encoding results of the DeepSC semantic communication model. Furthermore, we combine these encoding results with relevant information such as SINR to derive the relevant variables for semantic communication, eliminating additional data compression and storage steps originally required by the DeepSC model~\cite{ref58}.

\unskip  

%\begin{figure}[H]
%	\centering
%	\includegraphics[scale=0.4]{fig1.png}
%	\caption{Framework of semantic communication system}
%	\label{fig:fig1}
%\end{figure}   

\subsection{Scenario}\label{sect:s3dot1}
\begin{figure}[H]
	 
	\includegraphics[scale=0.4]{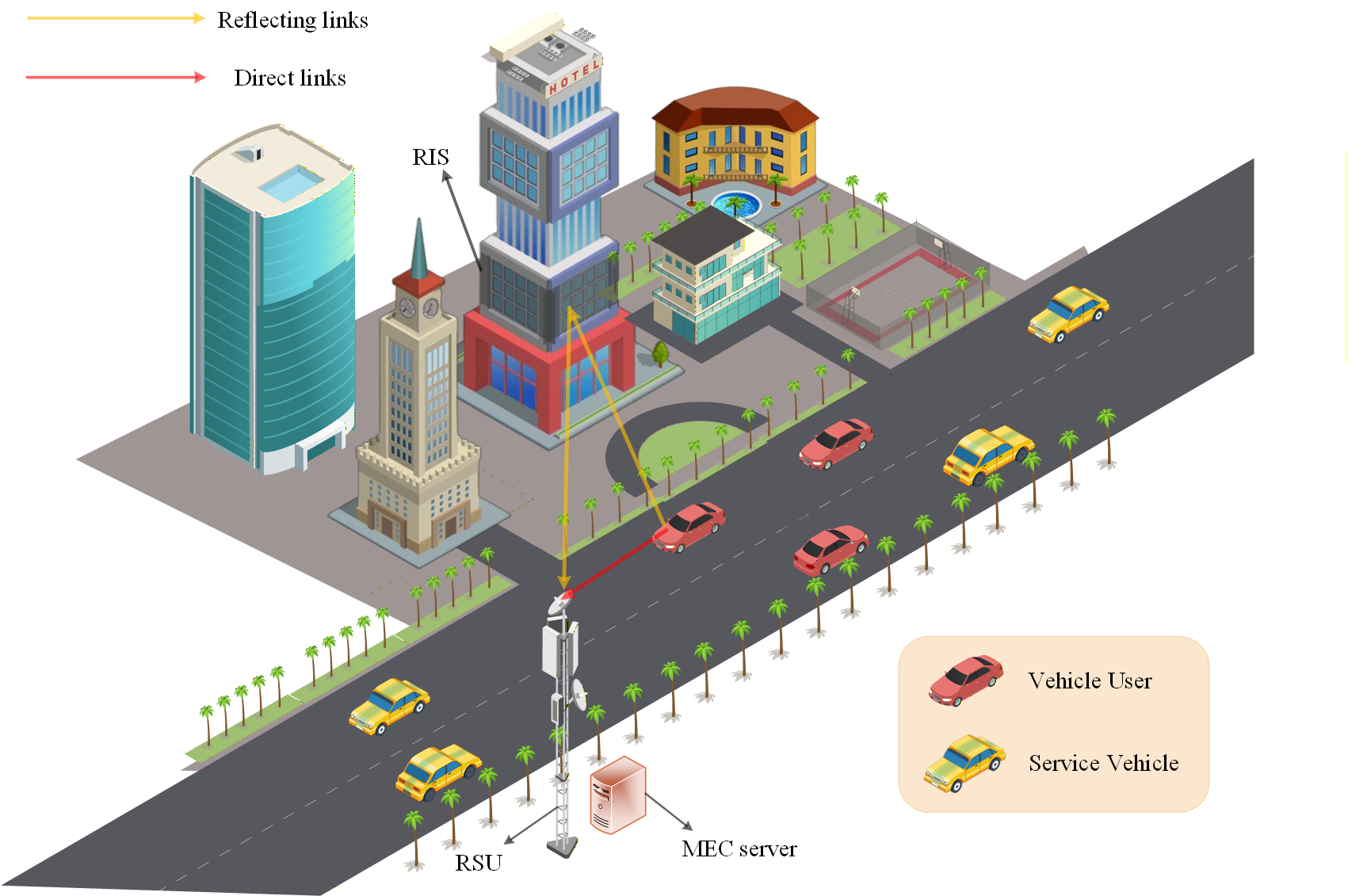}
	\caption{System model~diagram.}
	\label{fig:fig2}
\end{figure}

As shown in {Figure} \ref{fig:fig2}, the~RIS-aided semantic-aware VEC system comprises $K$ vehicle users $\mathcal{K} = \{1, \dots, K\}$ and edge nodes $\mathcal{J} = \{0, \dots, J\}$. Here, $0$ denotes the MEC-enabled RSU, while $1\sim J$ are SVs. Any vehicle other than your own may serve as an SV. Time is discretized into slots $\mathcal{T} = \{1, \dots, T\}$ of duration $\Delta t$. The~RSU is deployed in a relatively open area near the center and equipped with a large number of multiple-input multiple-output (MIMO) antennas. Due to the high vehicle mobility, each vehicle is equipped with an omnidirectional antenna. In~each time slot $\Delta t$, vehicles are uniformly distributed on the road. Regarding task characteristics, the~task arrivals of vehicle users follow a Poisson distribution, and~the text-based tasks generated by vehicle users have a fixed size. {In practical IoV scenarios, text-based tasks may vary in length. However, the~adopted DeepSC-based semantic encoder maps textual information into a controllable number of semantic symbols, which mitigates the impact of bit-level length variations. The~integration with DeepSC is realized by mapping the high-dimensional features extracted by the semantic encoder into a variable number of semantic symbols $v$. This allows the framework to adaptively perform 'semantic compression' based on the real-time channel quality $h_{k,j}^t$ and SINR $\gamma_{k,j}^t$. Specifically, we utilize the validated semantic parameter mapping from~\cite{yan} to ensure that the relationship between $v$, SINR, and~semantic similarity $\delta$ accurately reflects the performance of a physical DeepSC implementation. Furthermore, the~proposed two-tier hybrid framework exhibits inherent robustness: the PPO agent can adaptively tune the semantic symbol count $\nu_{k,j}^t$ based on real-time feedback, while the LP solver provides instantaneous optimal offloading ratios to compensate for the computational fluctuations, ensuring the system remains within the latency thresholds.} Each vehicle user adaptively partitions its task into three parts: local execution, V2I semantic offloading to the RSU, and~V2V semantic offloading to the SV. For~the target selection of V2V task offloading, a~vehicle user intends to select the nearest service~vehicle.

\subsection{RIS-Based Semantic Offloading~Model}\label{sect:s3dot2}
In our proposed model, we assume that within time slot $\Delta t$, the~position of the $k$-th vehicle user is $(x_k^t, y_k^t, z_k^t)$, the~position of the $j$-th edge node is $(x_j^t, y_j^t, z_j^t)$, and~the fixed-position RIS is located at $({\rm RIS}_x, {\rm RIS}_y, {\rm RIS}_z)$. Thus, we can obtain the channel gain between the $k$-th vehicle user and the $j$-th edge node: %MDPI: Please confirm if the format (italics or non-italics, bold or non-bold) in equations need to be retained and kept consistent in the whole doc.  The same to other variables with the same situation.
% I confirm that the format (italics/non-italics, bold/non-bold) of all variables in the equations has been checked and kept consistent throughout the entire manuscript.
\begin{equation}
	\mathbf{h}_{k,j}^t = \sqrt{l (d_{k,j}^t)^{-\eta_{k,j}}} \mathbf{g}_{k,j}^t, 
\end{equation}
where $l$ denotes the path loss at reference distance $d_0 = 1$ m; $d_{k,j}^t$ represents the distance between the $k$-th vehicle and $j$-th edge node, with~$\eta_{k,j}$ as the path loss exponent; and $\mathbf{g}_{k,j}^t$ accounts for the small-scale multipath~fading.

To mitigate the severe path loss during signal propagation, we deploy a large RIS on the surface of buildings adjacent to the road, as~illustrated in Figure~\ref{fig:fig2}, in~order to reconstruct the wireless propagation environment through intelligent reflection. Specifically, the~RIS consists of a large number of passive reflecting elements. In~addition, considering that the links between them are line-of-sight (LoS), the~channel between the two can be modeled as a static channel. Therefore, the~channel gain between the RIS and the $j$-th edge node can be obtained as follows:
\begin{equation}
	\mathbf{h}_{r,j} = \sqrt{l \left(d_{r,j}\right)^{-\eta_{r,j}} \sqrt{\frac{K}{1+K}} \mathbf{h}_{r,j}^{\text{LoS}}},
\end{equation}
where $d_{r,j}$ is the RIS-to-edge distance, $\eta_{r,j}$ is the path loss exponent, and~$K$ is the Rician factor. {Since both the RIS and the RSU are fixedly deployed in the urban IoV environment, this ensures the existence of a dominant LoS component. This provides a stable ``virtual LoS link'' to compensate for the severe multipath fading and occlusion commonly seen in direct V2I and V2V links.} The LoS component $\mathbf{h}_{r,j}^{\text{LoS}}$ is defined as follows: \begin{equation}
	\mathbf{h}_{r,j}^{\text{LoS}} = \left[1, e^{-j\frac{2\pi}{\lambda}D_r \sin\left(\theta_{r,j}\right)}, \dots, e^{-j\frac{2\pi}{\lambda}(N-1)D_r \sin\left(\theta_{r,j}\right)}\right]^T,
\end{equation}
 with carrier wavelength $\lambda$, element spacing $D_r$, departure angle $\theta_{r,j}$, and~total number of reflecting elements $N$. Similarly, the~channel gain between user $k$ and the RIS is:
\begin{equation}
	\mathbf{h}_{k,r}^t = \sqrt{l \left(d_{k,r}^t\right)^{-\eta_{k,r}} \sqrt{\frac{K}{1+K}} \mathbf{h}_{k,r}^{t,\text{LoS}}} \quad \forall k \in \mathcal{K}, \forall t \in \mathcal{T}. 
\end{equation}
Here, %MDPI: please check whether to keep the non-indent format, same below in highlight.
% I confirm.
 $d_{k,r}^t$ and $\eta_{k,r}$ denote the time-varying user-RIS distance and path loss exponent, respectively. The~LoS component is:  \begin{equation}
	\mathbf{h}_{k,r}^{t,\text{LoS}} = \left[1, e^{-j\frac{2\pi}{\lambda}D_r \sin\left(\theta_{k,r}^t\right)}, \dots, e^{-j\frac{2\pi}{\lambda}(N-1)D_r \sin\left(\theta_{k,r}^t\right)}\right]^T,
\end{equation}
 where $\theta_{k,r}^t$ denotes the angle of arrival (AoA) of the signal from the $k$-th vehicle user to the RIS at time slot $t$. 
 
 Based on the above RIS-assisted communication propagation characteristics, we further analyze the communication quality of the system. Assume that the system spectrum is divided into $B$ orthogonal resource blocks (RBs), denoted by the set $\mathcal{B} = \{1, 2, \dots, B\}$. {Vehicle-to-roadside link resources are scarce and show frequency selectivity. As~a result, multiple links may reuse the same RB and cause co-channel interference.} For any communication link $(k,j)$, if~it is allocated to  $b$-th resource block for transmission, i.e.,~$b \in \mathcal{B}$, SINR between the $k$-th vehicle user and the $j$-th edge node at time slot $t$ can be expressed as follows:
\begin{equation}
\gamma_{k,j}^{t,(b)} = \frac{p_{k,j} \left| \mathbf{h}_{k,j}^t +  \left( \mathbf{h}_{r,j}^{(m)} \right)^H \Theta^{t} \mathbf{h}_{k,r}^{t} \right|^2}{\sum_{(k',j') \neq (k,j)} \alpha_{k',j'}^{(b)} p_{k'} \left| \mathbf{h}_{k',j'}^{t} \right|^2 + \sigma^2}. 
\end{equation}
Here, $p_{k,j}$ denotes the transmission power from the $k$-th vehicle user to the $j$-th edge node, and~the term $\mathbf{h}_{k,j}^t +  \left( \mathbf{h}_{r,j}^{(m)} \right)^H \Theta^{t} \mathbf{h}_{k,r}^{t}$ represents the signal component reflected by the RIS. {We define the diagonal phase shift matrix of the RIS as $\Theta^t = \text{diag}\left( \mu_1 e^{j\theta_1^t}, \mu_2 e^{j\theta_2^t}, ..., \mu_N e^{j\theta_N^t} \right)$ for all $n \in [1,N]$, where $\mu_n \in [0,1]$ represents the transmission coefficient amplitude and phase shift for each element $n$.} Due to hardware constraints, the~phase shift can only be selected from a finite discrete set $\theta_n^t \in \Phi = \left\{ 0, \frac{2\pi}{2^q}, ..., \frac{2\pi(2^q-1)}{2^q} \right\}$, where $q$ is the control bit for the phase shift discretization degree. $\sum_{(k',j') \neq (k,j)} \alpha_{k',j'}^{(b)} p_{k',j'} \left| \mathbf{h}_{k',j'}^t \right|^2$ represents the aggregated interference from other links $(k',j')$ sharing the same resource block $b$, where $\alpha_{k',j'}^{(b)} \in \{0,1\}$ is a binary spectrum reuse variable (set to 1 if and only if link $(k',j')$ occupies resource block $b$). $\sigma^2$ is the additive white Gaussian noise~power.

Based on the analysis of the physical layer channel and SINR, we further construct a semantic-aware transmission model. This paper adopts DeepSC as the underlying text transmission framework. At~the transmitter, the~$k$-th vehicle user encodes its text task into a semantic symbol sequence using a DeepSC encoder %MDPI: Please state the name of the manufacturer, city, and country from where the equipment was sourced.
% The "DeepSC encoder" mentioned in the text is not a physical hardware device, but a deep learning-based software model for semantic encoding. It has been implemented via Python software based on the foundational framework proposed in reference [11] (Xie et al., 2021). Therefore, it does not have a physical manufacturer or geographic origin. To avoid ambiguity, I have replaced "DeepSC transmitter" with "DeepSC encoder" in the manuscript.
 and transfers it to the edge node $j$ (RSU or SV) over a wireless channel; the receiver decodes it using a DeepSC receiver to recover the original text information. Unlike traditional communication systems focusing on bit transmission rate, our system focuses on the effective transmission efficiency of semantic information. Therefore, we adopt the semantic rate as the core performance metric, with~the unit of semantic units per second (suts/s). At~time slot $t$, the~semantic rate $R_{k,j}^t$ between the $k$-th user and edge node $j$ is defined as follows:
\begin{equation}
R_{k,j}^t = \frac{W \cdot I_k}{L_k \cdot \nu_{k,j}^t} \delta_{k,j}^t,
\end{equation}
where $W$, $I_k$, and~$L_k$ denote the bandwidth, average semantic information, and~word length per sentence, respectively. The~semantic similarity $\delta_{k,j}^t \in [0,1]$ is a non-linear function of the number of semantic symbols $\nu_{k,j}^t$ and SINR $\gamma_{k,j}^{t,(b)}$, expressed as $\delta_{k,j}^t = f(\nu_{k,j}^t, \gamma_{k,j}^{t,(b)})$. {This mapping relationship $f(\cdot)$ is derived from the pre-trained DeepSC model. By~adopting the validated semantic parameter table from~\cite{yan}, we reuse the verified and reliable data, thus reducing the repetition of experimental work.}

{The semantic rate formula represents the effective throughput of information by coupling the physical layer bandwidth $W$ with the application layer's semantic richness. Unlike bit-rate, which treats all bits equally, this model prioritizes the 'semantic units'. By~optimizing $v_{k,j}^t$ as a decision variable, the~system can maintain a high semantic similarity $\delta_{th}$ even when RIS phase shifts $\Theta^t$ cannot fully mitigate deep fades.}
\subsection{Computational~Model}\label{sect:s3dot3}
{As illustrated in the three-path framework, the~total task volume $D_k$ generated by vehicle $k$ is split into three parallel components with offloading coefficients $\rho_k^{loc}$, $\rho_k^{RSU}$, and~$\rho_k^{SV}$. These paths operate concurrently, and~the end-to-end latency is determined by the bottleneck path. Each path's delay is modeled separately to reflect the distinct communication and computation characteristics of local, V2I, and~V2V resources.}
\begin{enumerate}[label=$\bullet$]
	\item Local computing: The local processing delay of tasks depends on the vehicle's computing capability. Denote $f_k^{\text{loc}}$ as local CPU frequency of $k$-th vehicle user, $C$ as CPU cycles per bit of data, and~$D_k$ as original bit data size. The~local computing delay $T_k^{\text{loc}}$ is expressed as follows:
\begin{equation}
		T_k^{\text{loc}} = \frac{\rho_k^{\text{loc}} D_k C}{f_k^{\text{loc}}}.
	\end{equation}
		
	\item RSU-assisted computing: Task offloading to RSU involves two delays: semantic data transmission delay and RSU computing delay. In~the transmission phase, DeepSC converts raw data to semantic symbol streams. A~conversion factor $H$ maps raw data $D_k$ to semantic task queue length (sentence count) $Q_k = D_k/H$. The~V2I link transmission delay $T_{k,0}^{\text{tr}}$ is:
\begin{equation}
		T_{k,0}^{\text{tr}} = \frac{\rho_k^{\text{RSU}} Q_k I_k}{R_{k,0}^t} = \frac{\rho_k^{\text{RSU}} Q_k \cdot (\nu_{k,0}^t L_k)}{W \cdot \delta_{k,0}^t}.
	\end{equation}
	In the computing phase, RSU evenly allocates  total computing capacity $F^{\text{RSU}}$, $U_0$ connected vehicles, and~thus the RSU processing delay $T_{k,0}^{\text{com}}$ is expressed as follows: \begin{equation}
		T_{k,0}^{\text{com}} = \frac{\rho_k^{\text{RSU}} D_k C}{F^{\text{RSU}}/U_0}.
	\end{equation}  
	Note that the return delay is neglected due to small computation results.
	\item SV-assisted computing: For V2V offloading, vehicles collaborate with the nearest SV. The~transmission delay $T_{k,j}^{\text{tr}}$ (where $j \neq 0$) is:
\begin{equation}
		T_{k,j}^{\text{tr}} = \frac{\rho_k^{\text{SV}} Q_k I_k}{R_{k,j}^t} = \frac{\rho_k^{\text{SV}} Q_k \cdot (\nu_{k,j}^t L_k)}{W \cdot \delta_{k,j}^t}, j \neq 0.
	\end{equation}
	 For computing resource allocation: $j$-th SV serves $U_j$ vehicles by sharing its computing capacity $F_j^{\text{SV}}$. A~service threshold $U_{\text{max}} \approx \lfloor K/J \rfloor$ is set to avoid overload while excessive requests are offloaded to the next nearest SV. Under~normal load, SV computing delay $T_{k,j}^{\text{com}}$ (where $j \neq 0$) is:
\begin{equation}
	 	T_{k,j}^{\text{com}} = \frac{\rho_k^{\text{SV}} D_k C}{F_j^{\text{SV}}/U_j}, j \neq 0.
	 \end{equation}
	 	 Return delay is neglected, consistent with V2I offloading.
\end{enumerate}

Since local execution, V2I offloading and V2V offloading run in parallel, the~total task delay of vehicle $k$ is the maximum of the three paths:
\begin{equation}
T_k^{\text{Total}} = \max\left\{ T_k^{\text{loc}}, T_{k,0}^{\text{tr}} + T_{k,0}^{\text{com}}, T_{k,j}^{\text{tr}} + T_{k,j}^{\text{com}} \right\}.
\end{equation}

\subsection{Problem~Formulation}\label{sect:s3dot4}
{   The aim of this is to improve %MDPI: please check there is content missing here.
% I have carefully checked this section and confirm that the subsection title is complete as intended, and there is no content missing in the text.
 VEC system efficiency by minimizing the overall delay. Specifically, we jointly optimize the offloading ratios, the~number of semantic symbols, and~the RIS phase shifts. The~resulting optimization problem, $\mathcal{P}1$, is defined as follows:}
\begin{equation} \label{eq:opt_constraints}
\begin{aligned}
	\mathcal{P}1: \quad &\min_{\substack{\rho_k^{\text{loc}}, \rho_k^{\text{RSU}}, \rho_k^{\text{SV}}, \nu_{k,j}^t,  \theta_n^t }} \sum_{k \in \mathcal{K}} T_k^{\text{Total}} \\
	\text{s.t.} \quad &C1: \ \rho_k^{\text{loc}}, \rho_k^{\text{RSU}}, \rho_k^{\text{SV}} \in [0,1], \ \forall k \\
	&C2: \ \rho_k^{\text{loc}} + \rho_k^{\text{RSU}} + \rho_k^{\text{SV}} = 1, \ \forall k \\
	&C3: \ \delta_{k,j}^t \geq \delta_{\text{th}}, \ \forall k,j,t \\
	&C4: \ \theta_n^t \in \Phi, \ \forall n \\
	&C5: \ \nu_{k,j}^t \in \{1, \dots, \nu_{k,j}^{\max}\} \\
	&C6: \ T_k^{\text{loc}} \leq T_k^{\text{max}}, \ \forall k \\
	&C7: \ T_{k,0}^{\text{tr}} + T_{k,0}^{\text{com}} \leq T_k^{\text{max}}, \ \forall k \\
	&C8: \ T_{k,j}^{\text{tr}} + T_{k,j}^{\text{com}} \leq T_k^{\text{max}}, \ \forall k,j, \ j \neq 0.
\end{aligned}
\end{equation}
In Problem $\mathcal{P}1$, Constraint C1 defines the range of task offloading proportions, while constraint C2 ensures the task is fully processed. Constraint C3 guarantees the semantic similarity of all links $(k,j)$ is no less than the threshold. Constraint C4 restricts RIS phase shifts to the feasible configuration set. Constraint C5 denotes that the number of semantic symbols is a finite discrete value. Finally, constraints C6, C7 and C8 ensure each task part meets the delay requirement, where $T_k^{\text{max}}$ denotes the maximum delay~threshold.

\section{Problem Formulation and~Solution} \label{sect:s4}
Since problem $\mathcal{P}1$ involves both continuous offloading ratio variables and the number of semantic symbols, as~well as RIS phase shift variables, the~objective function is non-convex. If~DRL is adopted to optimize them simultaneously, it will lead to an excessively large action space and state space, thereby prolonging the training time and increasing the uncertainty of the training results. To~address this, this paper adopts a two-layer collaborative hybrid solution framework. The~upper layer utilizes the PPO algorithm to learn the discrete policies for RIS phase shift selection and the number of semantic symbols selection. The~lower layer employs LP to achieve the optimal offloading ratio given the upper-layer actions, thus minimizing the end-to-end latency under the given RIS phase shifts and the number of semantic symbols choices. Specifically, to~effectively solve $\mathcal{P}1$, we decouple it into two mutually nested sub-problems and process them using this two-layer framework. First, given the number of semantic symbols and RIS phase shift variables, the~task offloading ratios are optimized to minimize the delay. Then, based on the feedback of the optimal task offloading ratios, the~optimal strategies for the number of semantic symbols and RIS phase shifts are~searched.

\subsection{PPO~Algorithm} \label{sect:s4dot1}
We employ PPO to optimize the discrete variables, i.e.,~the number of semantic symbols $v_{k,j}^t$ and phase shifts $\theta_n^t$. With~a given continuous offloading ratio $\rho$, $\mathcal{P}1$ is reformulated as sub-problem $\mathcal{P}2$:
\begin{equation}
	\begin{aligned}
		\mathcal{P}2: \quad &\min_{\substack{\nu_{k,j}^t, \theta_n^t \ }} \sum_{k \in \mathcal{K}} T_k^{\text{total}} \\
		\text{s.t.} \quad &C3, C4, C5, C7, C8.
	\end{aligned}
\end{equation}

This process is modeled as a Markov Decision Process (MDP), where the RSU agent interacts with the environment via state $s$, action $a$, and~reward $r$.

(1) %MDPI: we completed the parenthesis, please confirm; same as below.
% I have checked and confirmed that all the parentheses have been completed correctly. Thank you for the revision.
 The state at slot $t$ is:
\begin{equation}
s_t = \{L_t, \boldsymbol{\theta}_{k,r}^t, \boldsymbol{\theta}_n^t, \gamma_{k,j}^{t,(b)}\},
\end{equation}
	where $L_t$ denotes the coordinate positions of all vehicles, including the real-time 3D coordinates of vehicle users ($k \in \mathcal{K}$) and SVs ($j \in \mathcal{J}$); $\boldsymbol{\theta}_{k,r}^t$ represents the AoA of the signal from the $k$-th vehicle user to the RIS at time slot $t$; $\boldsymbol{\theta}_n^t$ denotes the phase shift coefficients of the RIS; and $\gamma_{k,j}^{t,(b)}$ is the SINR of the $b$-th RB at the current time~slot.

(2) Action space: The agent's action space is:
\begin{equation}
a_t = \{\nu_{k,j}^t, \boldsymbol{\theta}_n^t\},
\end{equation}
 where $\nu_{k,j}^t$ is the number of semantic symbols decision and $\theta_n^t$ is the discrete phase \mbox{shifts~decision.}

(3) Reward function: Upon executing action $a_t$, the~lower-layer LP solver computes the minimum system latency $T_k^{\text{total}}$. The~reward is defined as the negative total latency: \begin{equation}
	r_t = -T_k^{\text{total}}.
\end{equation}

We model the policy $\pi_\psi(a_t|s_t)$ as a Multivariate Gaussian Distribution to handle the high-dimensional action space efficiently. 
Although the optimization variables (RIS phase shifts and the number of semantic symbols) are inherently discrete, directly searching in a high-dimensional discrete space is computationally prohibitive. 
Therefore, we adopt a continuous relaxation strategy during the training phase. 
Specifically, the~Actor network outputs continuous actions within the range $[-1, 1]$, which are subsequently mapped to valid discrete values.
For a given state $s_t$, the~network outputs the mean vector $\mu(s_t)$ and maintains a standard deviation vector $\sigma$.
The continuous action sampling formula is:
\begin{equation} \label{19}
a_t = \tanh\left(\mu(s_t) + \sigma \odot z_t\right),\quad z_t \sim \mathcal{N}(0, I),
\end{equation}
where $\mathcal{N}(0, I)$ denotes the standard normal distribution, and~$\odot$ represents element-wise multiplication.
The $\tanh$ function constrains the raw actions $\boldsymbol{a}_t$ to the interval $[-1, 1]$.
To bridge the gap between the continuous output and discrete physical constraints, we perform the following mapping operations.
For the semantic symbol count $\nu_{k,j}^t$, the~continuous action $a_{\nu} \in [-1, 1]$ is linearly scaled and mapped to the nearest integer:
\begin{equation}
	\nu_{k,j}^t = \text{round}\left( \frac{a_{\nu} + 1}{2} \cdot (\nu_{max} - 1) \right) + 1.
\end{equation}
Similarly, for~the RIS phase shift $\theta_n^t$, the~action $a_{\theta} \in [-1, 1]$ is mapped to the nearest feasible discrete phase in set $\Phi$:
\begin{equation}
	\theta_n^t = \arg\min_{\phi \in \Phi} \left| \phi - \pi(a_{\theta} + 1) \right|.
\end{equation}
{The adoption of continuous relaxation followed by a rounding operation provides a practical balance between training efficiency and hardware constraints. By~enabling smooth gradient updates in a continuous action space, this strategy effectively circumvents the instability inherent in high-dimensional discrete optimization. }This hybrid approach leverages the efficient gradient updates of continuous PPO while ensuring that the executed actions strictly satisfy the discrete constraints of the physical system.
The Critic network $V_\phi(s_t)$ (parameterized by $\phi$) is used to estimate the value of the current state to assist in policy update. 
To reduce the variance in gradient estimation, the~algorithm uses the advantage function $A_t$ to evaluate the relative superiority of~actions.

The advantage function $A_t$ is calculated as the difference between the discounted cumulative return $G_t$ and the state value estimation:
\begin{equation} \label{20}
G_t = \sum_{l=0}^{T-t-1} \gamma^l r_{t+l},
\end{equation}
\begin{equation} \label{21}
A_t = G_t - V_\phi(s_t),
\end{equation}
where $\gamma$ is the discount factor. In~practical calculations, $A_t$ is usually standardized to further improve the numerical stability of training. The~core of the PPO algorithm lies in limiting the difference between the old and new policies. 
 The probability ratio $r_t(\psi)$ is defined as the ratio of the probability of the new policy $\pi_\psi$ to the old policy $\pi_{\psi_{old}}$ under the current action: $r_t(\psi) = \frac{\pi_\psi(a_t|s_t)}{\pi_{\psi_{old}}(a_t|s_t)}$. 
To prevent performance collapse caused by excessively large policy update steps, PPO introduces a clipping operation. The~loss function $L^{CLIP}(\psi)$ of the Actor is defined as follows:
\begin{equation}
L^{CLIP}(\psi) = \mathbb{E}_t\left[\min\left(r_t(\psi)A_t, clip\left(r_t(\psi), 1-\epsilon, 1+\epsilon\right)A_t\right)\right].
\end{equation}

In the formula, the~$clip(\cdot)$ function restricts the probability ratio to $[1-\epsilon, 1+\epsilon]$.
%here, $r_t = \frac{\pi_{\text{new}}(\mathcal{A}_t|\mathcal{S}_t)}{\pi_{\text{old}}(\mathcal{A}_t|\mathcal{S}_t)}$ denotes the probability ratio of the new and old policies, $\pi_{\text{new}}$ is the current policy, $\pi_{\text{old}}$ is the old policy used for data collection. The $\text{clip}(\cdot)$ function restricts $r_t$ to the interval $[1-\epsilon, 1+\epsilon]$, where $\epsilon$ is a clipping hyperparameter, this avoids policy oscillation caused by overly large single gradient updates, ensuring the algorithm's numerical stability. $A_t$ is the Generalized Advantage Estimation (GAE) value.
The Critic network is updated by minimizing the Mean Square Error (MSE) between the predicted value and the true return:
\begin{equation}
L^{VF}(\phi) = \mathbb{E}_t\left[\left(V_\phi(s_t) - G_t\right)^2\right].
\end{equation}

Considering the policy improvement, value fitting, and~exploration ability comprehensively, the~total loss function \(L^{total}\) is constructed as follows:
\begin{equation} \label{24}
L^{total} = -L^{CLIP}(\psi) + c_1 L^{VF}(\phi) - c_2 S[\pi_\psi](s_t),
\end{equation}
where \(c_1\) is the value loss coefficient; and \(c_2\) is the entropy coefficient; \(S[\pi_\psi]\) denotes the entropy of the policy distribution, which encourages the agent to maintain high randomness in the early stage of training and prevents premature convergence to local optima. Finally, by~using the Adam optimizer to minimize the above total loss function, the~algorithm can end-to-end learn the optimal RIS configuration and traffic steering~strategy.

\subsection{LP~Algorithm} \label{sect:s4dot2}
After the PPO agent outputs action $a_t$, semantic parameter and RIS properties are fixed, and~V2I, V2V link channels, transmission rates become constants; this removes the non-linear coupling between RIS configuration and offloading proportion in~P1.

We then use LP to optimize the continuous offloading proportion $\rho$ for minimal delay, transforming $\mathcal{P}1$ into $\mathcal{P}3$:
\begin{equation}
	\begin{aligned}
		\textbf{P3:} \quad &\min_{\rho} \sum_{k \in \mathcal{K}} T_k^{\text{total}} \\
		\text{s.t.} \quad &C1, C2, C6, C7, C8.
	\end{aligned}
\end{equation}

Based on the system's total latency, we first define the unit data processing time coefficients for the three paths: $\mu_k^{\text{loc}} = \frac{D_k C}{f_k^{\text{loc}}}, \quad \mu_k^{\text{RSU}} = \frac{Q_k I_k}{R_{k,0}(\mathcal{A}_t)} + \frac{D_k C}{F^{\text{RSU}}/U_0}, \quad \mu_k^{\text{SV}} = \frac{Q_k I_k}{R_{k,j}(\mathcal{A}_t)} + \frac{D_k C}{F_j^{\text{SV}}/U_j}$. 
Since constraints C6, C7, C8 relate to $\rho$, they can be simplified as follows:
\begin{equation}
\begin{aligned}
	\rho_k^{\text{loc}} \mu_k^{\text{loc}} - T_k^{\text{total}} &\le 0, \\
	\rho_k^{\text{RSU}} \mu_k^{\text{RSU}} - T_k^{\text{total}} &\le 0, \\
	\rho_k^{\text{SV}} \mu_k^{\text{SV}} - T_k^{\text{total}} &\le 0.
\end{aligned}
\end{equation} 

To solve $\mathcal{P}3$ efficiently with mature numerical optimization algorithms, we convert its physical model into the standard matrix-form optimization problem:
\begin{equation}
\begin{aligned}
	\min_{\boldsymbol{x}} \quad &\boldsymbol{c}^{\intercal}\boldsymbol{x} \\
	\text{s.t.} \quad &A\boldsymbol{x} \leq \boldsymbol{b}, \quad A_{\text{eq}}\boldsymbol{x} = \boldsymbol{b}_{\text{eq}}, \quad \boldsymbol{lb} \leq \boldsymbol{x} \leq \boldsymbol{ub}.
\end{aligned}
\end{equation}

We define the augmented optimization vector $\boldsymbol{x} \in \mathbb{R}^4$, which includes three offloading proportion variables and one delay auxiliary variable $\boldsymbol{x} = \left[ \rho_k^{\text{loc}}, \rho_k^{\text{RSU}}, \rho_k^{\text{SV}}, T_k \right]^{\intercal}$. The~goal is to minimize the fourth term of $\boldsymbol{x}$. Therefore, the~coefficient vector $\boldsymbol{c}$ is set as $\boldsymbol{c} = \left[ 0, 0, 0, 1 \right]^{\intercal}$.

Based on the simplified expressions of constraints C6, C7, C8, we construct the inequality constraint matrix $\mathbf{A}$ and vector $\boldsymbol{b}$:
\begin{equation}
\mathbf{A} = \begin{bmatrix}
	\mu_k^{\text{loc}} & 0 & 0 & -1 \\
	0 & \mu_k^{\text{RSU}} & 0 & -1 \\
	0 & 0 & \mu_k^{\text{SV}} & -1
\end{bmatrix}, \quad \boldsymbol{b} = \begin{bmatrix} 0 \\ 0 \\ 0 \end{bmatrix}.
\end{equation}

Constraint C2 requires the sum of offloading proportions to be 1, corresponding to the equality constraint matrix and vector:
\begin{equation}
A_{\text{eq}} = \left[ 1, 1, 1, 0 \right], \quad \boldsymbol{b}_{\text{eq}} = \left[ 1 \right].
\end{equation}
Constraint C1 and the maximum delay threshold $T_k^{\text{max}}$ form the variable bounds:
\begin{equation}
 \boldsymbol{lb} = \left[ 0, 0, 0, 0 \right]^{\intercal}, \quad \boldsymbol{ub} = \left[ 1, 1, 1, T_k^{\text{max}} \right]^{\intercal}.
 \end{equation}

Through the above transformation, the~problem is reformulated as a standard convex optimization problem, which is solved using the interior-point method to obtain the globally optimal solution $\boldsymbol{x}^*$. The~minimum system total delay $T_{\text{sys}} = \sum_{k \in \mathcal{K}} T_k^{\text{Total}}$ is calculated based on $\boldsymbol{x}^*$, which is fed back to the upper-layer PPO algorithm as a reward signal to guide the update of the number of semantic symbols and RIS configuration~policy.

{The lower-layer LP solver plays a critical role in mitigating the impact of task size fluctuations and queueing effects. For~each time slot $t$, the~LP optimizer receives the actual task demand $D_k^t$ and the reshaped channel gains $h_{k,j}^t$ from the upper-layer RIS configuration. By~formulating the sub-problem as a standard linear program with $4K$ variables, including auxiliary delay variables $T_k$, the~solver yields the optimal offloading ratios $\rho_k^{loc}, \rho_k^{RSU}, \rho_k^{SV}$ instantaneously. If~a vehicle experiences a sudden increase in task volume or a signal blockage due to mobility, the~LP solver automatically re-balances the workload among the three available computational paths. This dynamic compensation mechanism ensures that even under non-ideal conditions such as service vehicle movement or transient traffic bursts, the~end-to-end delay is minimized and kept within the safety-critical threshold $T_k^{max}$.}
\subsection{Algorithm Implementation and Training~Phase} \label{sect:4.3}
Based on the constructed loss function and optimization objectives, we designed a complete training process for the PPO algorithm to achieve the joint optimization of RIS configuration, the~number of semantic symbols and task~scheduling.

The training process follows the online reinforcement learning paradigm. In~the interaction phase, the~Actor network first outputs the action movement towards RIS phase shifts and the number of semantic symbols at current state. Subsequently, the~optimizer within the environment reconstructs the channel environment for this action decision and solves the LP problem to obtain the optimal relay decision and minimum transmission delay, which are then fed back as reward signals. In~the model update phase, trajectory data is sampled periodically from the experience replay buffer to calculate the advantage function, and~the PPO clipped objective function constraint strategy is used for parameter update. This approach not only ensures the training stability but also coordinates the optimization of physical layer resource allocation and hyperparameter selection in the semantic layer, thus ultimately maximizing the long-term cumulative reward of the system. The~complete execution steps of the two-layer collaborative optimization solution framework proposed in this paper are presented in Algorithm \ref{alg:ppo_training}.

\begin{algorithm}[H]
	\caption{PPO Algorithm Training~Phase}
	\label{alg:ppo_training}
	\begin{algorithmic}[1] % [1] 表示显示行号
		\REQUIRE 
		Maximum training episodes $E_{max}$, 
		Maximum time steps per episode $T_{max}$, 
		Network parameter update interval $T_{update}$;
		PPO hyperparameters: learning rate $\alpha$, clipping coefficient $\epsilon$, discount factor $\gamma$
		\ENSURE 
		Trained Actor network parameters $\psi^*$, 
		Trained Critic network parameters $\phi^*$
		
		\STATE Randomly initialize the weight parameters of Actor network and Critic network.
		\STATE Initialize the experience replay buffer $\mathcal{D}$ for storing trajectory data. Initialize the global time step counter $t_{step} = 0$.
		
		\FOR{$episode = 1$ \TO $E_{max}$}
		\STATE Reset the environment and observe the initial state $s_1$.
		\FOR{$t = 1$ \TO $T_{max}$}
		\STATE Input the current state $s_t$ into the Actor network. Generate the joint action $a_t$ via Equation \eqref{19}.
		\STATE Call the LP solver to obtain the optimal task offloading ratio $p^*$ and the corresponding minimum system delay $t_{total}$.
		\STATE Calculate the reward $r_t$ and get the next state $s_{t+1}$.
		\STATE Store the transition tuple $(s_t, a_t, r_t, s_{t+1})$ in the experience buffer $\mathcal{D}$.
		\STATE Update the current state $s_t \leftarrow s_{t+1}$ and increment the time step counter $t_{step}$.
		
		\IF{$t_{step} > T_{update}$}
		\STATE Based on the buffer $\mathcal{D}$, calculate the discounted cumulative return and advantage function via Equations \eqref{20} and %MDPI: we merged the citaitons, please confirm.
		% I have checked and confirmed that all citations have been merged correctly. Thank you for the revision.
 \eqref{21}.
		
		\FOR{$k = 1$ \TO $K$}
		\STATE Calculate the probability ratio of the new and old policies.
		\STATE Calculate the total loss function via Equation \eqref{24}.
		\STATE Minimize the total loss using Adam optimizer and update the parameters of Actor and Critic networks.
		\ENDFOR
		\ENDIF
		\ENDFOR
		\ENDFOR
	\end{algorithmic}
\end{algorithm}

\subsection{Testing~Phase} \label{sect:4.4}
In the testing phase, operations such as network parameter adjustments and gradient updates are not required. Instead, the~trained model is directly employed to obtain the optimal scheme for RIS configurations and task offloading ratios. Note that each test round consists of 200 test time slots, and~the final result is derived from the average performance across all time slots within that round. The~pseudo-code for the testing phase is shown in Algorithm \ref{alg:ppo_testing}.

\begin{algorithm}[htbp]
	\caption{PPO Algorithm Testing~Phase}
	\label{alg:ppo_testing}
	\begin{algorithmic}[1]
		\REQUIRE Trained Actor network parameters $\psi^*$, number of test episodes $E_{test}$, maximum time steps per episode $T_{max}$
		\ENSURE Average system delay
		
		\FOR{$episode = 1$ \TO $E_{test}$}
		\STATE Reset the environment and obtain the initial state $s_1$.
		\FOR{$t = 1$ \TO $T_{max}$}
		\STATE Input the current state $s_t$ into the Actor network.
		\STATE Turn off exploration noise and output the deterministic action $a_t = \tanh(\mu(s_t))$.
		\STATE Parse the action $a_t$ to get RIS phase shifts and the number of semantic symbols.
		\STATE Solve Problem P3 to obtain the optimal unloading ratio $\rho^*$ and the minimum system delay $T_{sys}^*$.
		\STATE Observe the next state $s_{t+1}$.
		\STATE Update the state $s_t \leftarrow s_{t+1}$.
		\ENDFOR
		\ENDFOR
	\end{algorithmic}
\end{algorithm}
\vspace{-\baselineskip} % 抵消空行
{\subsection{Complexity~Analysis}}
{To substantiate the practical feasibility of the proposed two-layer optimization framework, a~rigorous analysis of the computational complexity is indispensable. The~overall system complexity per time slot is characterized by the synergy between the upper-layer PPO inference and the lower-layer LP optimization.}

{The computational overhead of the upper-layer decision-making predominantly stems from the forward propagation of the Actor network during the online inference phase. Assuming the Actor network comprises $L$ fully connected layers with $n_l$ neurons in the $l$-th layer, the~complexity of processing a single state input is defined as $\mathcal{O}\left(\sum_{l=1}^{L-1} n_l \cdot n_{l+1}\right)$. We denote this neural network inference complexity as $NN_{inference}$. Given that the network architecture—characterized by the number of layers and neurons—is fixed upon the completion of offline training and is typically maintained at a compact scale, $NN_{inference}$ represents a constant and minimal computational demand. This ensures that the vehicle users, even with limited onboard processing power, can generate optimal RIS phase shifts $\theta_{n}^{t}$ and semantic symbol configurations $\nu_{k,j}^{t}$ in near real time.}
	
{Following the PPO-based discrete decision-making, lower-layer optimization involves solving a standard linear programming problem to determine the continuous offloading ratios $\rho_{k}^{loc}, \rho_{k}^{RSU}$, and~$\rho_{k}^{SV}$. By~formulating this sub-problem as a standard LP form with $4K$ variables (including three offloading proportions and one auxiliary delay variable for each of the $K$ vehicles), we employ the interior-point method for resolution. The~computation complexity of this stage is approximately $\mathcal{O}(K^{3.5})$. Consequently, the~total computation complexity of the proposed framework for each time step is expressed as $\mathcal{O}(NN_{inference} + K^{3.5})$. This decoupled architecture effectively circumvents the exponential complexity incurred by high-dimensional non-convex joint optimization, thereby guaranteeing the low-latency requirements of safety-critical vehicular communications.}

\section{Simulation Results and~Analysis}
\label{sect:s5}
Simulations are implemented via Python 3.8 and PyTorch 1.12.1 %MDPI: Please state the version number of the software or the accessed website link and its accessed date.
% The version number of the software has been indicated.
 in an urban scenario where an RIS is deployed to assist vehicles suffering from severe attenuation. The~setup comprises 1 RSU at ($-$10, 150, 25), 1 RIS at (10, 175, 25). Each vehicle's speed is fixed at 20 m/s and assumed to remain constant thereafter, which is used for theoretical modeling and training setup. In~actual simulation, vehicle speed remains constant while inter-vehicle spacing correlates with vehicle count. Vehicle users generate a task load of 0.4 Mbit per time slot (generated via Poisson distribution). Key simulation environment parameters are shown in  \tabref{tabref:table-1}. {These parameters are carefully selected to align with the 5G-V2X communication standards and practical vehicular hardware specifications. Specifically, the~semantic communication settings are based on the pre-trained DeepSC model to ensure compatibility, while the computing capacities of RSUs and SVs are set to reflect typical edge server performance in urban IoV scenarios.}

\begin{table}[H] 
	\tablesize{\small}
	\caption{Simulation environment parameter~table.}
	\label{tabref:table-1}
	\renewcommand{\arraystretch}{1.1} % 调整行间距（1.5倍，可按需修改）
	\newcolumntype{C}{>{\centering\arraybackslash}X}
	\begin{tabularx}{\textwidth}{CCCC}
		\toprule
		\textbf{Parameters} & \textbf{Value} & \textbf{Parameters} & \textbf{Value}\\
		\midrule
		$\eta_{r,j}$        & 2.2            & $p_{k,j}$           & 0.2~W\\
		$N$        & $6 \times 6$   & $\sigma^2$          & $1.44 \times 10^{-10}$ W\\
		$K$                 & 15             & $\eta_{k,j}$        & 3.5\\
		$\eta_{k,r}$        & 2.2            & $W$                 & 360~kHz\\
		$C$                 & 1000 cycles/bit& $I_k$               & 100\\
		$f_k^{\text{loc}}$  & 2 GHz          & $F^{\text{RSU}}$    & 6~GHz\\
		$F_j^{\text{SV}}$   & 2 GHz          & $L_k$               & 20\\
		$H$                 & 1200 bit       & $\delta_{\text{th}}$& 0.9\\
		$\alpha$            & \mbox{0.0003 (actor)/0.001 (critic)}   & $\gamma$            & 0.6\\
		$\epsilon$          & 0.2            & $E_{\text{max}}$& 5000\\
		\bottomrule
	\end{tabularx}
\end{table}
We compare our proposed PPO method with the following methods:

\begin{itemize}
	\item Genetic Algorithm (GA)~\cite{ref59}: GA simulates natural selection and genetic mechanisms, searching for the global optimal solution within the solution space through selection, crossover, and~mutation~operations.
	
	\item Quantum-behaved Particle Swarm Optimization (QPSO): QPSO is a heuristic benchmark adopting the QPSO algorithm~\cite{ref60}. Compared with the standard Particle Swarm Optimization (PSO) algorithm, QPSO eliminates the velocity vector and utilizes wave functions from quantum mechanics to describe the motion state of particles. It has been proven to possess stronger global search capabilities and fewer control parameters.
\end{itemize}

\begin{figure}[H]
 
	\includegraphics[scale=0.75]{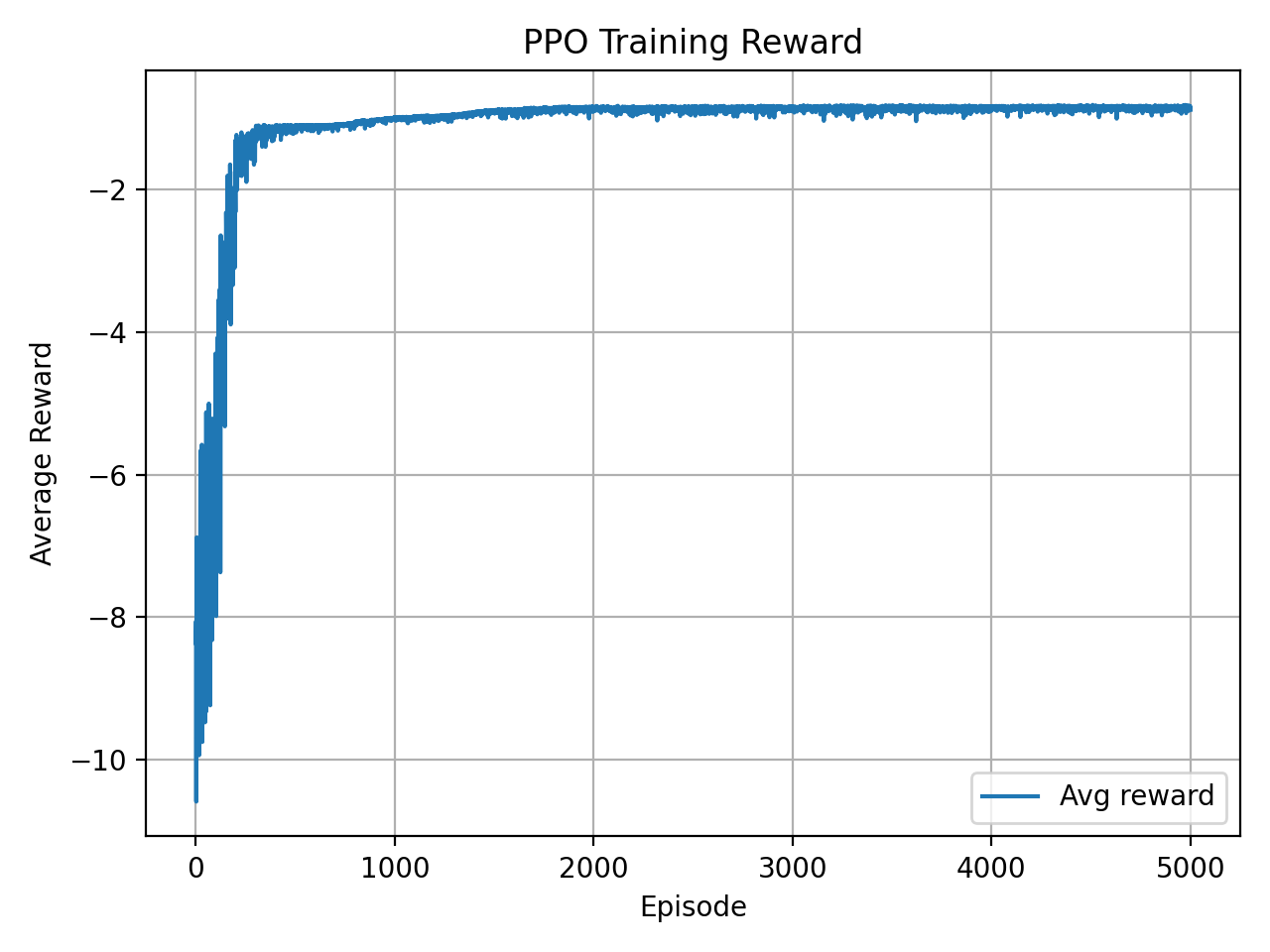}
	\caption{Reward convergence plot of~PPO.}
	\label{reward_curve}
\end{figure}

{Figure}~\ref{reward_curve} illustrates the trend of average cumulative reward of the proposed algorithm during the training process. In~this figure, the horizontal axis represents the number of training episodes, while the vertical axis denotes the average system reward achieved within each episode. It can be observed that the training process exhibits a distinct convergence trend, demonstrating favorable convergence characteristics and learning efficiency of PPO~algorithm.

\begin{figure}[H]
	 
	\includegraphics[scale=0.6]{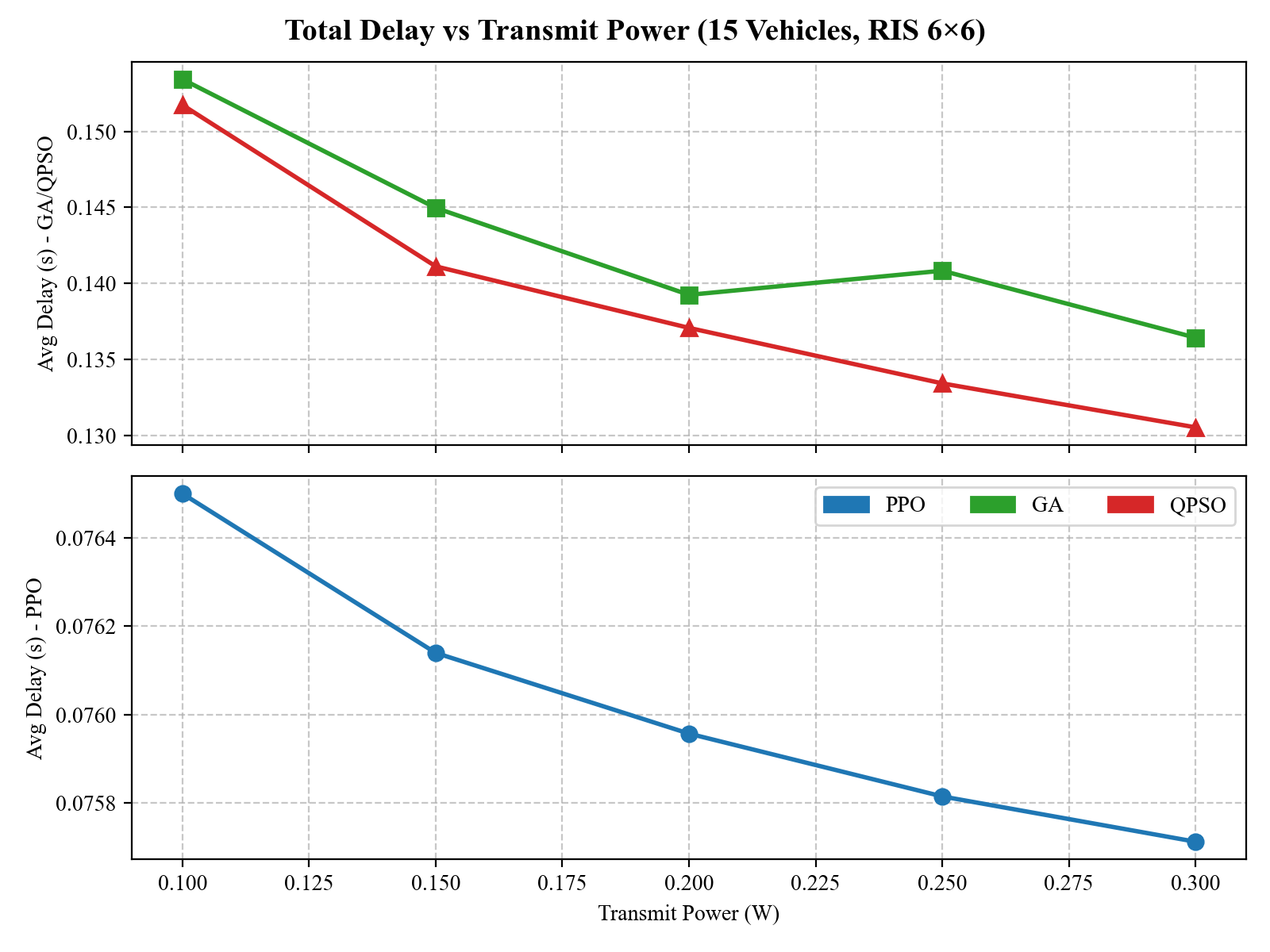}
	\caption{Performance comparison of average delay under varying vehicle transmission~powers.}
	\label{analysis_delay_vs_power}
\end{figure}

{Figure}~\ref{analysis_delay_vs_power} depicts the impact of vehicle transmit power on the average total system delay under the experimental setting of 15 vehicles and an RIS size of $6 \times 6$. As~illustrated, with~the vehicle transmit power increasing from 0.1 W to 0.3 W, the~system delay of all algorithms monotonically decreases. This is attributed to the fact that increased transmit power effectively improves the SINR at the receiver, thereby enhancing channel capacity and reducing data transmission latency. Notably, the~proposed PPO algorithm significantly outperforms GA and QPSO benchmark algorithms across the entire power range. Specifically, the~proposed algorithm achieves a performance improvement of approximately 40\% to 50\%. {Traditional heuristic algorithms often fluctuate and get trapped in local optima. In~contrast, the~PPO algorithm generates a smoother and continuously descending curve.} This fully validates the robustness of the DRL framework in handling high-dimensional continuous action spaces, as~well as its superiority in maximizing system efficacy through precise control of RIS phase shifts and the number of semantic~symbols.

\vspace{-6pt}

\begin{figure}[H]
	 
	\hspace{-20pt}\includegraphics[scale=0.6]{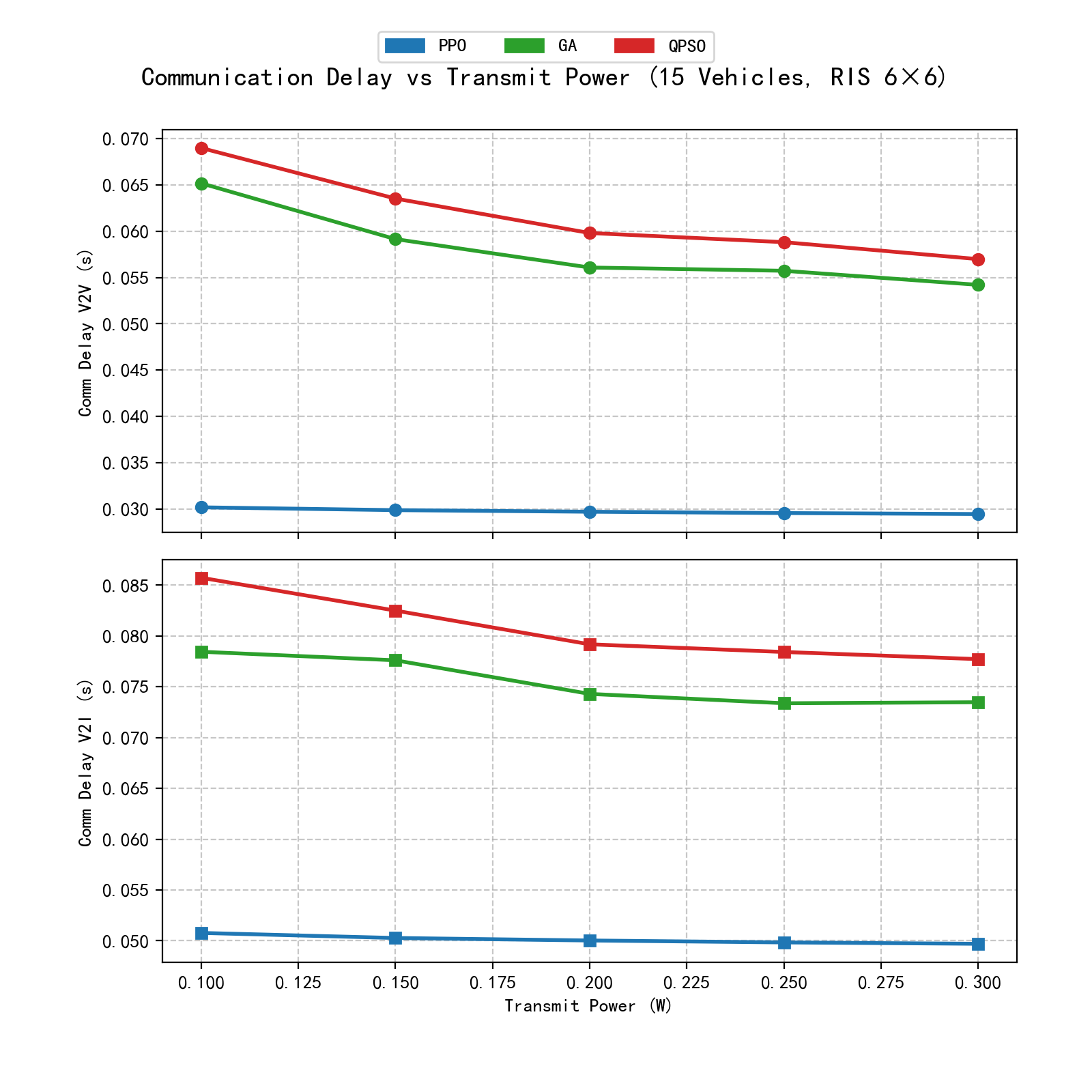}
	\caption{Average delay of V2V and V2I links versus vehicle transmission~power.}
	\label{analysis_comm_delay_vs_power}
\end{figure}

\newpage
{Figure}~\ref{analysis_comm_delay_vs_power} analyzes the communication delay components for V2V and V2I links under different transmit powers. When the transmit power increases from 0.1 W to 0.3 W, SINR is also improved accordingly, thus leading to reduced delays for all algorithms on both links. The~proposed PPO algorithm achieves the lowest delay, especially in the V2V link, where it stabilizes at around 0.030 s, providing over 45\% performance gain. For~the V2I link, PPO also maintains the best performance. These results confirm that the PPO agent effectively mitigates co-channel interference caused by spectrum reuse while enhancing the desired signal strength, thus satisfying the low-latency requirements of~IoV.

\begin{figure}[H]
	 
	\hspace{-2pt}\includegraphics[scale=0.5]{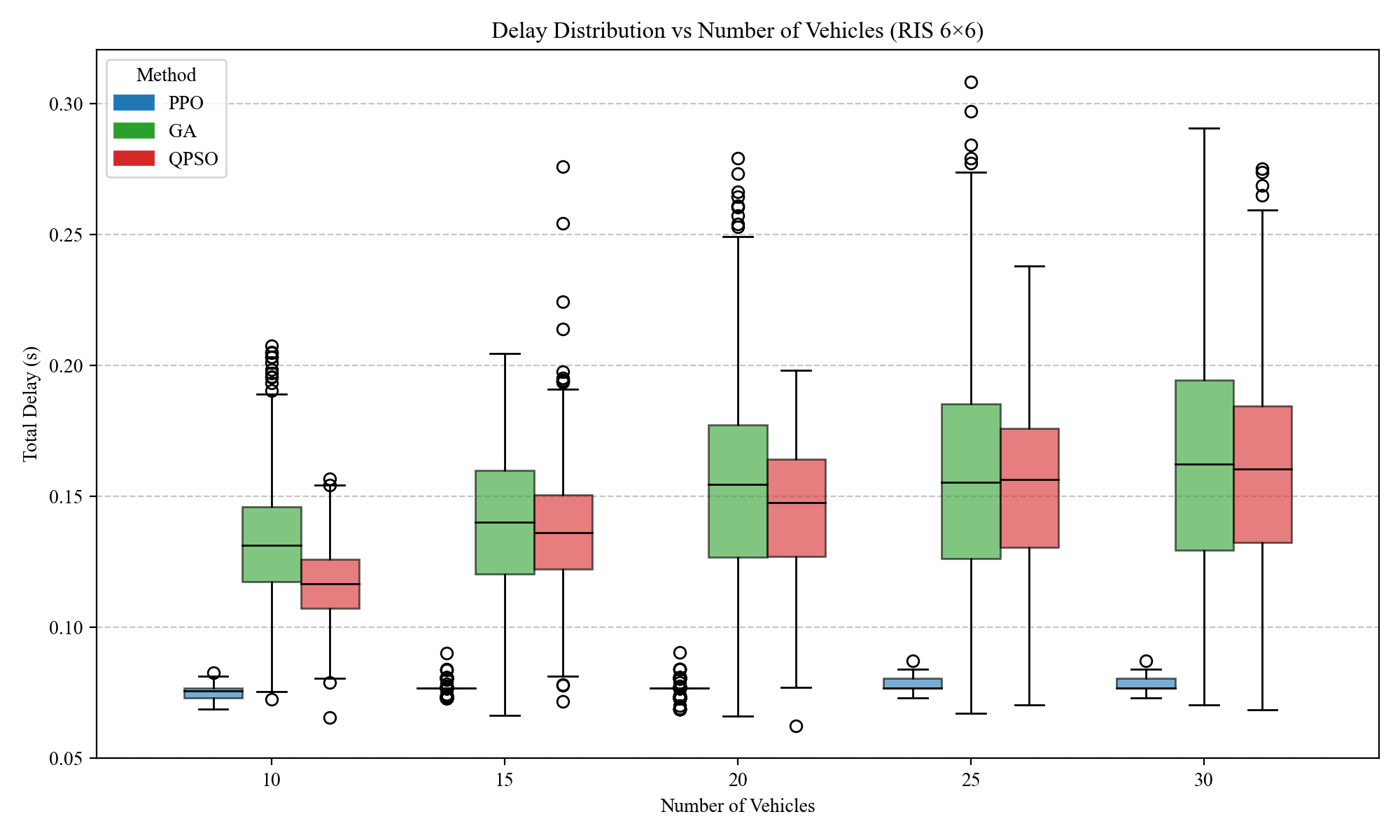}
	\caption{Boxplot %MDPI: The color are not matched with the explanation, please unify them if possible.
	% I have carefully checked the figure and confirmed that the colors in the plot are consistent with the legend explanation, and no further unification is needed.
 of average delay under different numbers of~vehicles.}
	\label{boxplot_delay_vs_vehicles}
\end{figure}

{Figure}~\ref{boxplot_delay_vs_vehicles} utilizes boxplots to evaluate algorithm stability with varying vehicle densities. The~outliers beyond the Interquartile Range (IQR) highlight instances where algorithms are trapped in poor local optima. {When the vehicle count reaches 30, GA and QPSO show elongated boxes and frequent outliers. This indicates significant instability and stochastic fluctuations in high-dimensional spaces.} In contrast, PPO maintains a compact distribution with minimal outliers, demonstrating its robustness in preventing extreme latency and ensuring fairness during~congestion.

\begin{figure}[H]
	 
	\includegraphics[scale=0.6]{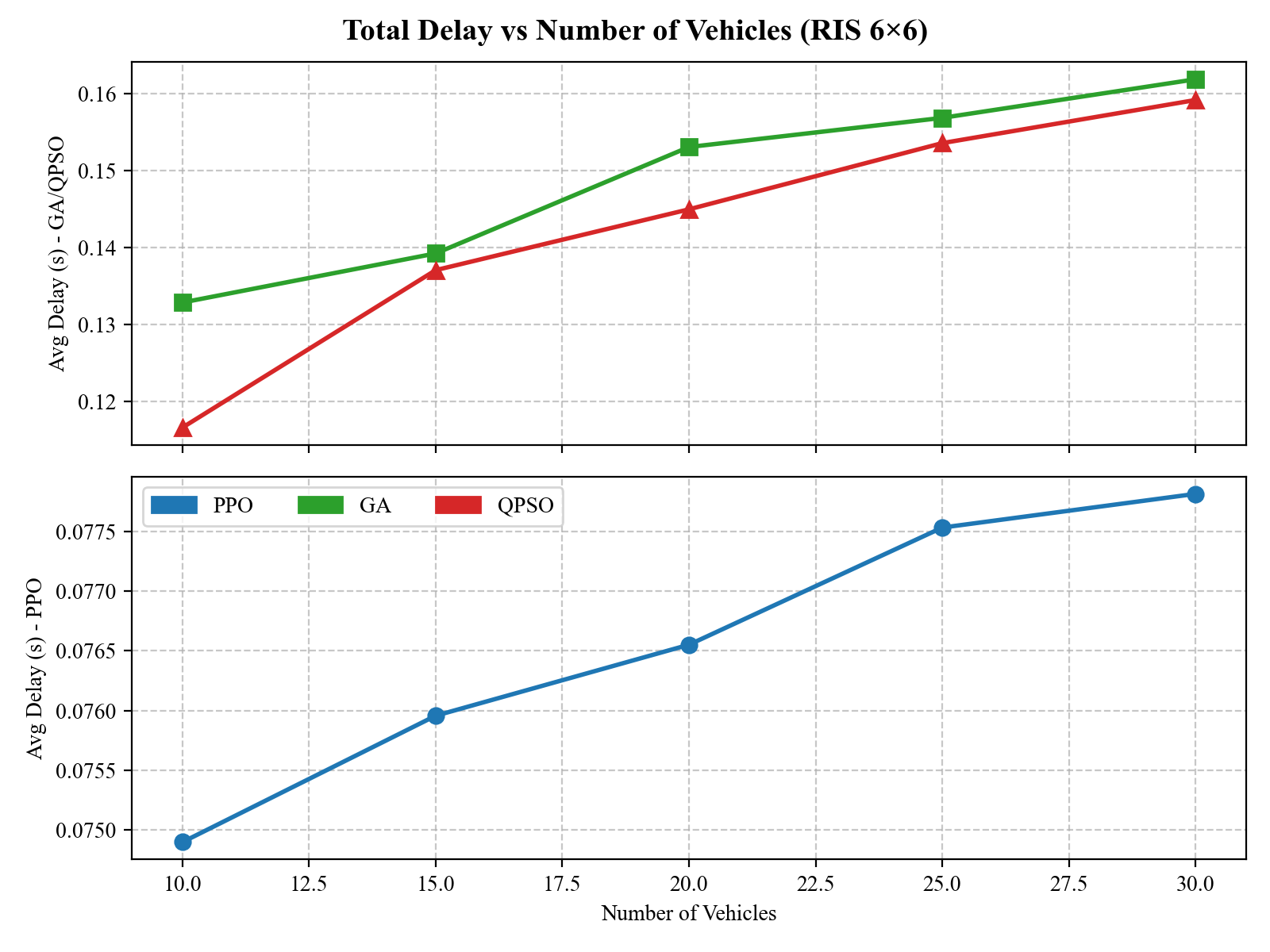}
	\caption{Average delay versus number of~vehicles.}
	\label{analysis_delay_vs_vehicles}
\end{figure}

{Figure}~\ref{analysis_delay_vs_vehicles} illustrates the system delay under dynamic vehicle densities. As~observed, increasing vehicle density exacerbates spectrum competition and co-channel interference, leading to higher delays across all schemes. However, the~proposed PPO algorithm demonstrates superior scalability compared to the benchmarks. Even in a congested scenario with 30 vehicles, PPO maintains a low delay with a minimal growth rate, whereas GA and QPSO exhibit rapid performance degradation. This confirms the robustness of the DRL-based approach in terms of resource allocation within interference-limited~environments.

\begin{figure}[H]
	 
	\includegraphics[scale=0.6]{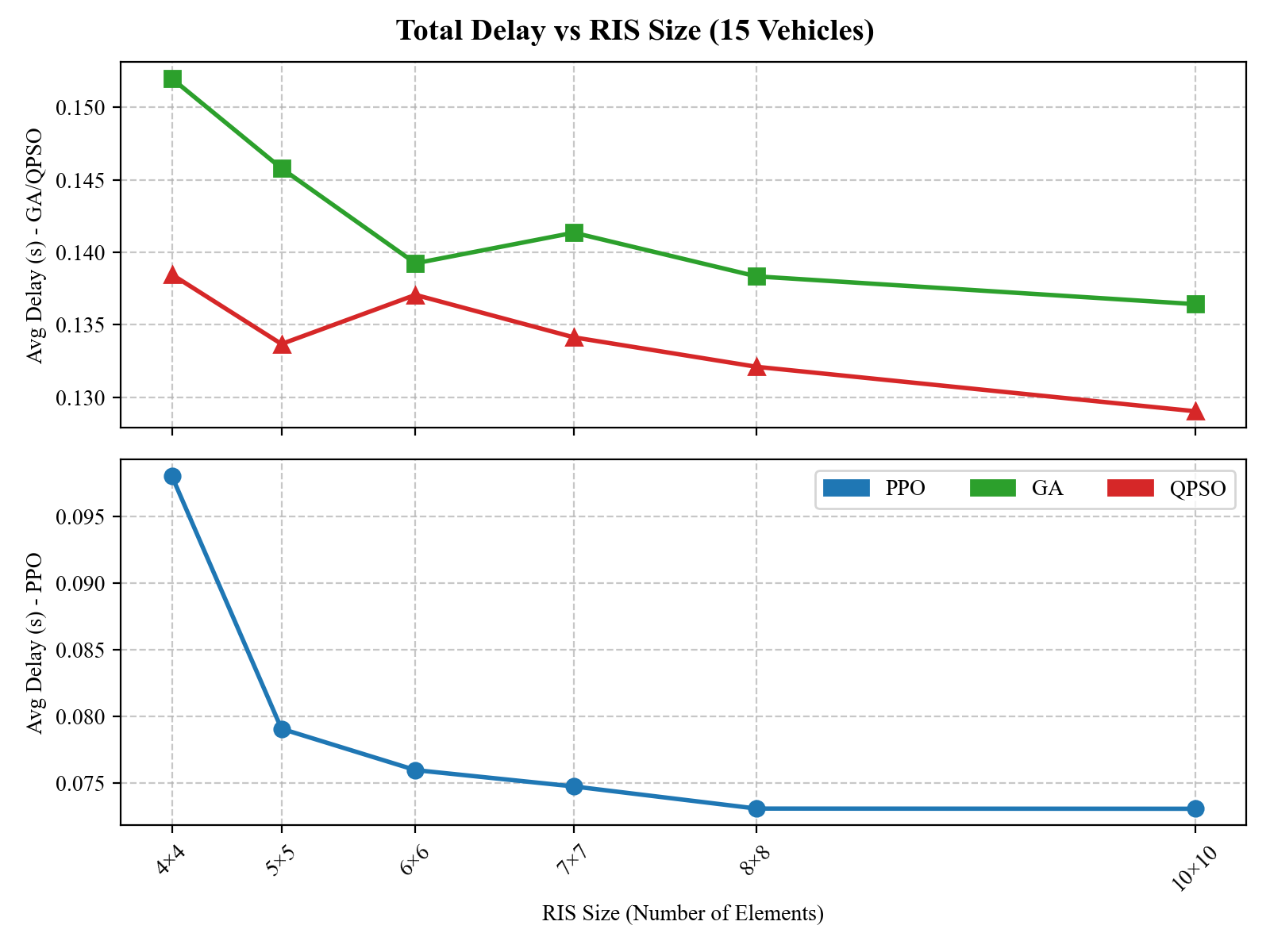}
	\caption{Average delay versus number of RIS-reflecting~elements.}
	\label{analysis_delay_vs_ris}
\end{figure}

{Figure}~\ref{analysis_delay_vs_ris} illustrates the impact of RIS scale on system delay. As~the number of elements increases from $4 \times 4$ to $10 \times 10$, system delay decreases for all schemes due to enhanced passive beamforming gains. However, PPO consistently outperforms GA and QPSO, which struggle with the ``curse of dimensionality'' in large-scale phase optimization. This demonstrates PPO's superior ability to handle high-dimensional continuous action spaces and fully exploit large-scale RIS~hardware.
   
\begin{figure}[H]
	 
	\includegraphics[scale=0.5]{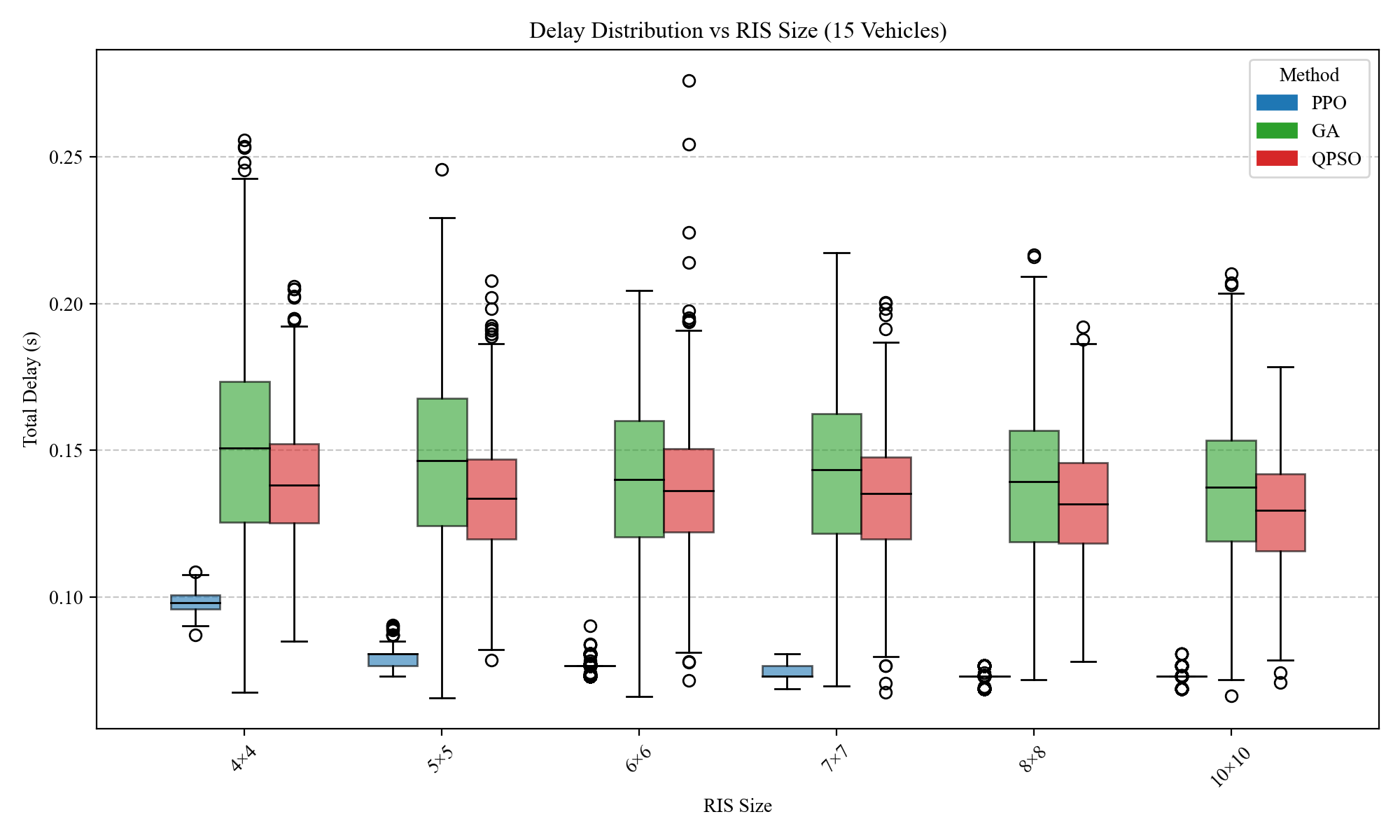}
	\caption{Boxplot %MDPI: The color are not matched with the explanation, please unify them if possible.
	% I have carefully checked the figure and confirmed that the colors in the plot are consistent with the legend explanation, and no further unification is needed.
 of total delay under different numbers of RIS-reflecting~elements.}
	\label{boxplot_delay_vs_ris}
\end{figure}

{Figure}~\ref{boxplot_delay_vs_ris} employs boxplots to evaluate the algorithm stability with varying RIS scales. The~expanding Interquartile Ranges (IQRs) and frequent outliers observed in GA and QPSO indicate severe instability and entrapment in local optima as dimensionality increases. Conversely, PPO exhibits compact distributions with minimal outliers, demonstrating its robustness in overcoming the ``curse of dimensionality'' associated with large-scale RIS~optimization.

{In summary, the~simulation results with varying vehicle densities and RIS scales provide a comprehensive ablation analysis of the proposed framework. The~RIS component is essential for reconfiguring the physical channel, as~evidenced by the latency reduction observed when the number of reflecting elements increases. The~semantic communication module enables semantic compression, which is the key driver behind the 40–50\% performance improvement over traditional bit-level optimization schemes~\cite{refzheng}. Finally, the~two-tier hybrid architecture (PPO-LP) ensures computation efficiency and stability. By~delegating the continuous offloading task to the LP solver, the~PPO agent can effectively handle the complex discrete search space of RIS phase shifts without succumbing to the curse of dimensionality.}

\section{Conclusions} \label{sect:s6}
In this paper, we have proposed a novel RIS-assisted semantic-aware VEC framework to tackle the conflicting requirements of ultra-low latency and high reliability in complex dynamic vehicular networks.
By integrating link-level RIS enhancement with semantic feature transmission, we constructed a robust three-path offloading model that significantly mitigates channel fading and reduces transmission overhead.
To solve the formulated non-convex joint optimization problem involving coupled discrete and continuous variables, we devised a two-tier hybrid solution framework.
Specifically, we employed a continuous relaxation-based PPO algorithm to efficiently search for optimal discrete configurations of RIS phase shifts and the number of semantic symbols while utilizing LP to optimize continuous offloading ratios.
Extensive simulation results demonstrate that the proposed approach outperforms varying baselines, reducing the average system latency by approximately 40\% to 50\% as compared to that observed with the GA and QPSO algorithms.
Crucially, our method exhibits superior robustness and scalability in high-density scenarios, which effectively overcomes the ``curse of dimensionality'' in large-scale RIS optimization.
Future work will focus on investigating the updating mechanisms of a dynamic semantic knowledge base and exploring multi-RIS cooperative coverage to further enhance system adaptability in wider~areas.

%%%%%%%%%%%%%%%%%%%%%%%%%%%%%%%%%%%%%%%%%%

%%%%%%%%%%%%%%%%%%%%%%%%%%%%%%%%%%%%%%%%%%
\vspace{6pt} 

%%%%%%%%%%%%%%%%%%%%%%%%%%%%%%%%%%%%%%%%%%
%% optional
%\supplementary{The following supporting information can be downloaded at:  \linksupplementary{s1}, Figure S1: title; Table S1: title; Video S1: title.}


% Only used for preprtints:
% \supplementary{The following supporting information can be downloaded at the website of this paper posted on \href{https://www.preprints.org/}{Preprints.org}.}

%\begin{tabularx}{\textwidth}{lll}
%\toprule
%\textbf{Name} & \textbf{Type} & \textbf{Description} \\
%\midrule
%S1 & Python script (.py) & Script of python source code used in XX \\
%S2 & Text (.txt) & Script of modelling code used to make Figure X \\
%S3 & Text (.txt) & Raw data from experiment X \\
%S4 & Video (.mp4) & Video demonstrating the hardware in use \\
%\bottomrule
%\end{tabularx}
%}

%%%%%%%%%%%%%%%%%%%%%%%%%%%%%%%%%%%%%%%%%%
\authorcontributions{Conceptualization, W.F.; methodology, J.Z. and W.F.; software, J.Z. and W.F.; validation, Q.W. and Q.F.; formal analysis, P.F. and W.F.; investigation, J.Z.; resources, W.F. and Q.W.; data Curation, W.F.; writing---original draft, W.F. and J.Z.; writing---review and editing, W.F., Q.W. and P.F.; visualization, J.Z.; supervision, W.F. and Q.W.; project administration and funding acquisition, Q.W. All authors have read and agreed to the published version of the manuscript.}

\funding{This research was funded %MDPI: Information regarding the funder and the funding number should be provided. Please check the accuracy of funding data and any other information carefully.
% All funding data and related information have been carefully verified and confirmed to be accurate.
 in part by the
	National Natural Science Foundation of China under Grant 61701197; in part
	by the Basic Research Program of Jiangsu under Grant BK20252084; in part by
	the National Key Research and Development Program of China under Grant
	2021YFA1000500(4) and in part by the 111 Project under Grant~B23008.}

\institutionalreview{Not~applicable.}

\informedconsent{Not applicable.}

\dataavailability{Data are disclosed via GitHub (\url{https://github.com/qiongwu86/PPO-Based-Hybrid-Optimization-for-RIS-Assisted-Semantic-Vehicular-Edge-Computing.git} \textls[-25]{(accessed on 24 February 2026)}.)%MDPI: Please add the access date (format: Date Month Year), e.g., accessed on 1 January 2020.)
% I have added the access date in the required format.
}

\conflictsofinterest{Author Qiang Fan was employed by the company Qualcomm. The remaining authors declare that the research was conducted in the absence of any commercial or financial relationships that could be construed as a potential conflict of interest. %MDPI: we noticed that aff 3 is a company. so we revised the statenment according to our rules, please check and confirm.
% I have checked the revised statement regarding Affiliation 3 and confirm that it is accurate and in compliance with the journal rules.
} 

%%%%%%%%%%%%%%%%%%%%%%%%%%%%%%%%%%%%%%%%%%
%% Optional

%% Only for journal Encyclopedia
%\entrylink{The Link to this entry published on the encyclopedia platform.}

%%%%%%%%%%%%%%%%%%%%%%%%%%%%%%%%%%%%%%%%%%
%\isPreprints{}{% This command is only used for ``preprints''.
\begin{adjustwidth}{-\extralength}{0cm}

\reftitle{References %MDPI: %Dear authors,
%Please do not delete nonredundant references or add any new references. According to our requirements, no content changes are allowed after the article is accepted. Hope you can understand.
%Please review all the references to ensure that there are no retractions. If any are found, remove them or replace them with new ones.
% I have carefully checked all references and confirm that no references have been deleted or newly added. No retracted articles are included.
}

\PublishersNote{}
%\isPreprints{}{% This command is only used for ``preprints''.
\end{adjustwidth}
\end{document}